\documentclass[10pt,twocolumn,twoside] {IEEEtran}

\usepackage{graphicx,amsfonts}
\usepackage{subfig}
\usepackage[usenames,dvipsnames]{color}
\usepackage{cite}
\ifCLASSINFOpdf
\else
\fi
\usepackage[cmex10]{amsmath}
\usepackage{array,multirow}
\usepackage{mdwmath}
\usepackage{mdwtab}
\usepackage{algorithm}
\usepackage{algcompatible}
\begin{document}

\title{A State-Space Approach to Dynamic Nonnegative Matrix Factorization}

\author{Nasser Mohammadiha, Paris Smaragdis, Ghazaleh Panahandeh, Simon Doclo
\thanks{
Published on IEEE Trans. Signal Process. on Dec. 2014. Link to the original paper: http://ieeexplore.ieee.org/document/6996052/

Citation: N. Mohammadiha, P. Smaragdis, G. Panahandeh, S. Doclo, "A State-Space Approach to Dynamic Nonnegative Matrix Factorization", IEEE Trans. Signal Process., vol. 63, no. 4, pp. 949--959, Dec. 2014.}}
%


\maketitle

\begin{abstract}
Nonnegative matrix factorization (NMF) has been actively investigated and used in a wide range of problems in the past decade. A significant amount of attention has been given to develop NMF algorithms that are suitable to model time series with strong temporal dependencies. In this paper, we propose a novel state-space approach to perform dynamic NMF (D-NMF). In the proposed probabilistic framework, the NMF coefficients act as the state variables and their dynamics are modeled using a multi-lag nonnegative vector autoregressive (N-VAR) model within the process equation. We use expectation maximization and propose a maximum-likelihood estimation framework to estimate the basis matrix and the N-VAR model parameters. Interestingly, the N-VAR model parameters are obtained by simply applying NMF. Moreover, we derive a maximum a posteriori estimate of the state variables (i.e., the NMF coefficients) that is based on a prediction step and an update step, similarly to the Kalman filter. We illustrate the benefits of the proposed approach using different numerical simulations where D-NMF significantly outperforms its static counterpart. Experimental results for three different applications show that the proposed approach outperforms two state-of-the-art NMF approaches that exploit temporal dependencies, namely a nonnegative hidden Markov model and a frame stacking approach, while it requires less memory and computational power.
\end{abstract}


\IEEEpeerreviewmaketitle

\section{Introduction\label{sec:Introduction}}

Nonnegative matrix factorization (NMF) \cite{Lee2000} is an approach
to obtain a low-rank representation of nonnegative data. In NMF,
a nonnegative data matrix $\mathbf{X}$ is factorized into a product of
two nonnegative matrices $\mathbf{W}$ and $\mathbf{H}$ such that
$\mathbf{W}\mathbf{H}$ provides a good approximation of $\mathbf{X}$ with respect to (w.r.t.) a predefined criterion.
In our notation, each column of $\mathbf{X}$ corresponds to a multivariate observation in time.
$\mathbf{W}$ and $\mathbf{H}$ are referred to as the basis matrix and NMF coefficient matrix, respectively, where each row of $\mathbf{H}$ represents the activity of its corresponding
basis vector over time.

In many signal processing applications, e.g., audio processing and
analysis of time series, the consecutive columns of $\mathbf{X}$
exhibit a strong temporal correlation. In the basic NMF approach, however, each
observation is treated individually. A simple and useful approach to alleviate this problem is to stack the consecutive columns of the data matrix into high-dimensional super-vectors, and to apply NMF to learn high-dimensional basis vectors. This frame stacking approach is one of the key ingredients in so-called exemplar-based representations \cite{Gemmeke2011a}. More rigorously, to model the temporal dependencies
in NMF, three main approaches have been developed in the past: 1)
Convolutive NMF \cite{Smaragdis2007,Wang2009}, in which the dependencies
are usually imposed on the basis matrix $\mathbf{W}$. 2) Smooth NMF \cite[and references therein]{Fevotte2009,Mohammadiha2013d}
where each row of $\mathbf{H}$ is individually constrained to evolve smoothly over time. 3) Approaches that combine NMF and hidden Markov
model (HMM) paradigms \cite{Mysore2010,Nakano2010,Mohammadiha2013c,Mohammadiha2013}.
In the so-called nonnegative hidden Markov model (N-HMM) \cite{Mysore2010,Mohammadiha2013c},
the NMF coefficient vectors are assumed to be sparse
and a first-order Markovian chain is considered over the index of
the active coefficients. These approaches are explained and compared in \cite{Smaragdis2014} in a unified framework.

Kalman filtering and its nonlinear extensions \cite{Kailath2000,Simon2006}
have been extensively studied within the signal processing community
to exploit temporal correlations in an optimal way. The basic
Kalman filter is based on a linear state-space model in which both
the process noise and observation noise are assumed to be Gaussian-distributed.
The goal of the Kalman filter is then to find a minimum-mean-square-error (MMSE) estimator of the state variables given the current and
past observations. This estimator is obtained by minimizing the Bayesian
mean square error (MSE) where no additional constraints are imposed
on the state variables. If the noise distributions are not Gaussian,
the Kalman filter still provides the optimal linear MMSE estimator \cite{Kailath2000}.

Recently, there has been some research on developing Kalman filters
subject to state constraints. In addition to the model reparameterization,
the projection and pseudo-observation approaches are two usual solutions to handle constraints \cite{Julier2007,Simon2010,Carmi2010}. In the projection
approach, the unconstrained estimate (after the observation update) is projected
to satisfy the constraints. In the pseudo-observation approach, however,
a fictitious observation is considered using which the unconstrained
estimate is further updated similarly to a real observation update.
For example, in \cite{Carmi2010} Kalman filtering with sparseness constraints on the state
variables is addressed where a pseudo-observation
approach is developed. In this paper, we focus on dynamic filtering in
a state-space representation where both the state variables and the observations
are considered to be nonnegative. In this case, the mentioned approaches can not be
used to optimally apply the constraints. This is not only because
the state variables are nonnegative, but also the distribution of
the nonnegative observations may be far from a Gaussian distribution.

Very recently, nonnegative dynamical systems \cite{Mohammadiha2013b,Fevotte2013} have been proposed in which the NMF coefficients act as the state variables. Compared
to N-HMM, the continuous state-spaces utilized in these approaches provide
a richer mechanism to benefit from the temporal dependencies. To use the temporal dynamics in the estimation of the NMF coefficients, we proposed a two-step algorithm in \cite{Mohammadiha2013b} based on a prediction and an update step. However, none of the estimators for the NMF coefficients and the dynamic model parameters were optimally derived for the specified assumptions. In the current work, we further refine the theoretical foundations of our previous study in \cite{Mohammadiha2013b} and derive new estimators and present new examples and results.

In this paper, we formulate the dynamic NMF using a novel state-space representation and use the expectation-maximization (EM) algorithm to derive optimal update rules for the NMF coefficients and the model parameters.
We consider a probabilistic formulation of NMF \cite{Smaragdis2006}
and develop a state-space representation to model the temporal dependencies
in NMF. The process equation, which describes the evolution of the
NMF coefficients, is based on an exponential distribution whose parameter
is given by a multi-lag nonnegative vector autoregressive (N-VAR) model. The
observation equation is similar to static NMF where the
observations are assumed to be drawn from a multinomial distribution.
The choice of these distributions is based on both their appropriateness to model nonnegative data and the possibility
to derive closed-form solutions. We propose a maximum a posteriori
(MAP) approach to estimate the state variables $\mathbf{H}$. The
obtained MAP estimate that consists of a prediction step and an update step is a filtering solution since it is only conditioned
on the current and past observations. Additionally, we derive maximum
likelihood (ML) estimates of the basis vectors $\mathbf{W}$ and the
N-VAR model parameters. We show that the ML estimate of the N-VAR model parameters is obtained by simply applying NMF, which is well suited to
our nonnegative framework. We provide numerical simulations for three
examples, i.e., tracking the frequency of a single sinusoid in
noise, separation of two sources with similar basis matrices, and speaker-dependent and -independent speech denoising examples. We compare the performance of the proposed D-NMF approach to the performance of the static NMF approach, the N-HMM in \cite{Mysore2010}, and the frame stacking approach in \cite{Gemmeke2011a}. Our simulations show that the D-NMF approach outperforms these competing algorithms, while it is less complex and hence it is a better choice for real-time applications.

The remainder of the paper is organized as follows: Section \ref{sec:Nonnegative-Matrix-Factorization}
provides a short overview of NMF. The proposed dynamic NMF using a state-space model is presented in
Section \ref{sec:Proposed-Dynamic-NMF}. Numerical simulations for several problems
are presented in Section \ref{sec:Numerical-Simulations}.

\section{Nonnegative Matrix Factorization\label{sec:Nonnegative-Matrix-Factorization}}
Nonnegative Matrix Factorization is a method using which
a $K\times T$-dimensional nonnegative matrix $\mathbf{X}=\left\{ x_{kt}\right\} $ is approximated
as $\mathbf{W}\mathbf{H}$, where the $K\times I$-dimensional matrix $\mathbf{W}=\left\{ w_{ki}\right\} $
and the $I\times T$-dimensional matrix $\mathbf{H}=\left\{ h_{it}\right\} $ are both constrained
to be nonnegative. The model order $I$, i.e., the number of columns in $\mathbf{W}$,
is usually less than $K$, i.e., the number of rows in $\mathbf{X}$, such that
a dimension reduction is also achieved using NMF. The $t$-th columns of $\mathbf{X}$ and $\mathbf{H}$ are denoted by $\mathbf{x}_t$ and $\mathbf{h}_t$, respectively. The nonnegativeness
property is usually helpful to interpret the factorization using the
underlying physical phenomena. In the deterministic NMF approaches \cite{Lee2000}, a cost function measuring the approximation error is minimized under the given nonnegativity constraints.
Popular choices for the cost function include Euclidean distance
(in EUC-NMF), Kullback-Leibler divergence (in KL-NMF), and the Itakura-Saito
divergence (in IS-NMF) \cite{Lee2000,Fevotte2009}.

In the following, we briefly describe the IS-NMF since we will use it
in our algorithm. Letting $\hat{\mathbf{X}}=\mathbf{W}\mathbf{H}$, the IS
divergence for the NMF problem is defined as:
\begin{equation}
d_{\text{IS}}\left(\mathbf{X\|}\hat{\mathbf{X}}\right)=\sum_{k=1}^{K}\sum_{t=1}^{T}\left(\frac{x_{kt}}{\hat{x}_{kt}}-\log\frac{x_{kt}}{\hat{x}_{kt}}-1\right).\label{eq:IS_cost}
\end{equation}

A widely used approach to minimize NMF cost functions is using
the multiplicative update rules, which minimize the cost function in
an iterative manner. For the IS-NMF, these update rules are given
as (see, e.g., \cite{Fevotte2009}):
\begin{eqnarray}
\mathbf{H} & \leftarrow & \mathbf{H}\odot\frac{\mathbf{W}^{\top}\left(\left(\mathbf{W}\mathbf{H}\right)^{-2}\odot\mathbf{X}\right)}{\mathbf{W}^{\top}\left(\mathbf{W}\mathbf{H}\right)^{-1}},
\label{eq:updateH_IS} \\
 \mathbf{W} & \leftarrow & \mathbf{W}\odot\frac{\left(\left(\mathbf{W}\mathbf{H}\right)^{-2}\odot\mathbf{X}\right)\mathbf{H}^{\top}}{\left(\mathbf{W}\mathbf{H}\right)^{-1}\mathbf{H}^{\top}},
\label{eq:updateW_IS}
\end{eqnarray}
where $\top$ denotes matrix transpose, $\odot$ represents element-wise multiplication, and the division and powers are performed element-wise.
These updates are iteratively performed until a local minimum of the
cost function is found.

In contrast to the deterministic NMF approaches, probabilistic formulations
facilitate deriving the desired estimates in a statically optimal
manner. In the next section, we present our D-NMF algorithm that is
based on probabilistic latent component analysis (PLCA)\cite{Smaragdis2006}.
\section{Proposed Dynamic NMF\label{sec:Proposed-Dynamic-NMF}}

\subsection{Statistical Model Description\label{sub:Statistical-Model-Description}}

We propose a state-space approach to perform dynamic nonnegative factorization.
In this approach, the NMF coefficients $\mathbf{h}_t$ are assumed to evolve over
time according to the following nonnegative vector autoregressive (N-VAR) model:
\begin{multline}
\left\{ \begin{array}{ll}
f\left(\mathbf{h}_{t}\mid\mathbf{A},\mathbf{h}_{t-1},\ldots\mathbf{h}_{t-J}\right)=\text{v-exp}\left(\mathbf{h}_{t};\sum_{j=1}^{J}\mathbf{A}_{j}\mathbf{h}_{t-j}\right) \vspace{2mm} ,\\
f\left(\mathbf{x}_{t}\mid\mathbf{W},\mathbf{h}_{t}\right)=\text{mult}\left(\mathbf{x}_{t};\gamma_{t},\mathbf{W}\mathbf{h}_{t}\right),
\end{array}\right.\label{eq:state-space-rep}
\end{multline}
 where $f(\cdot)$ denotes a probability density function, $\mathbf{x}_{t}$ denotes the nonnegative observation vector at time $t$ with $\gamma_{t}=\sum_{k}x_{kt}$, $J$ is the order of the N-VAR model, $\mathbf{A}_j$ denotes the $I\times I$-dimensional N-VAR model parameters corresponding to the $j$-th lag (and $\mathbf{A}$ denotes the union of $\mathbf{A}_j,\forall j$), $\text{v-exp}\left(\mathbf{h}_{t};\boldsymbol{\eta}_{t}\right)$
is the exponential probability density function over the vector $\mathbf{h}_{t}$
with independent elements $h_{it}$ as:
\begin{eqnarray}
\text{v-exp}\left(\mathbf{h}_{t};\boldsymbol{\eta}_{t}\right) & = & \prod_{i=1}^{I}\eta_{it}^{-1}e^{-h_{it}/\eta_{it}},\label{eq:exp_prior}
\end{eqnarray}
and $\text{mult}\left(\mathbf{x}_{t};\gamma_{t},\mathbf{p}_{t}\right)$
represents a multinomial distribution as:
\begin{eqnarray}
\text{mult}\left(\mathbf{x}_{t};\gamma_{t},\mathbf{p}_{t}\right) & = & \gamma_t!\prod_{k=1}^{K}\frac{p_{kt}^{x_{kt}}}{x_{kt}!},\label{eq:mult_dist}
\end{eqnarray}
where $!$ denotes the factorial and $\sum_{k}p_{kt}=1$. The conditional expected values of $\mathbf{h}_{t}$ and $\mathbf{x}_{t}$ under the model (\ref{eq:state-space-rep}) are given by:
\begin{multline}
\left\{ \begin{array}{ll}
E\left(\mathbf{h}_{t}\mid\mathbf{A},\mathbf{h}_{t-1},\ldots\mathbf{h}_{t-J}\right)=\sum_{j=1}^{J}\mathbf{A}_{j}\mathbf{h}_{t-j} \vspace{2mm} ,\\
E\left(\mathbf{x}_{t}\mid\mathbf{W},\mathbf{h}_{t}\right)=\left( \sum_{k}x_{kt} \right)\mathbf{W}\mathbf{h}_{t} \vspace{2mm},
\end{array}\right.\label{eq:expected_values}
\end{multline}
which is used to obtain an NMF approximation of the input data as $\mathbf{x}_{t}\approx (\sum_{k}x_{kt})\mathbf{W}\mathbf{h}_{t}$.

The distributions in \eqref{eq:state-space-rep} are chosen to be appropriate for nonnegative data. For example, it is well known that the conjugate prior for the multinomial likelihood
is the Dirichlet distribution. However, it can be shown that the obtained
state estimates in this case are no longer guaranteed to be nonnegative.
Therefore, we propose to use the exponential distribution in (\ref{eq:state-space-rep})
for which, as will be shown in Section \ref{sub:Estimation-Algorithm}, the obtained state estimates
are always nonnegative. In addition, a closed-form solution can be derived under the given statistical assumptions, see Section \ref{sub:Estimation-Algorithm}.

If we discard the first equation in (\ref{eq:state-space-rep}), we
recover the basic PLCA algorithm \cite{Smaragdis2006}. This special case (corresponding to $J=0$) is referred to as the static NMF as it does not model temporal dependencies. Here, the observations $\mathbf{x}_{t}$ are assumed
to be count data over $K$ possible categories. Using the PLCA notation,
each vector $\mathbf{h}_{t}$ is a probability vector that represents
the contribution of each basis vector in explaining the observation,
i.e., $h_{it}=f\left(z_{t}=i\right)$ where $z_{t}$ is a latent variable
used to index the basis vectors at time $t$. Moreover, each column
of $\mathbf{W}$ is a probability vector that contains the underlying
structure of the observations given the latent variable $z$ and is referred to as a basis vector. More
precisely, $w_{ki}$ is the probability that the $k$-th element of
$\mathbf{x}_t$ will be chosen in a single draw from the multinomial
distribution in (\ref{eq:state-space-rep}), i.e., $w_{ki}=f\left(\mathbf{x}_t=\mathbf{e}_{k}\mid z_t=i\right)$
with $\mathbf{e}_{k}$ being a $K$-dimensional indicator vector whose $k$-th
element is equal to one (see \cite{Mohammadiha2013b} for more explanation). Note that (by definition) $w_{ki}$ is time-invariant. In the following, this notation is
abbreviated to $w_{ki}=f\left(k\mid z_t=i\right)$.

It is worthwhile to compare (\ref{eq:state-space-rep}) to the state-space
model utilized in the Kalman filter and to highlight the main differences between the two. First, all the variables
are constrained to be nonnegative in (\ref{eq:state-space-rep}).
Second, the process and observation noises are embedded into the
specified distributions, which is different from the additive Gaussian
noise utilized in the Kalman filtering. Finally, in the process equation,
we have used a multi-lag N-VAR model. In our proposed algorithm, different lags can have different importance weights, which will be discussed in Section \ref{sub:Estimation-Algorithm}.
It is also important to note that we aim to estimate both state-space model parameters ($\mathbf{W}$ and $\mathbf{A}$) and state variables $\mathbf{H}$, where Kalman filter only estimates $\mathbf{H}$, given a priori determined $\mathbf{W}$ and $\mathbf{A}$.

In Section \ref{sub:Estimation-Algorithm}, we derive an expectation\textendash{}maximization
(EM) algorithm to compute maximum likelihood (ML) estimates of $\mathbf{A}$
and $\mathbf{W}$ and to compute a MAP estimate of the state variables
$\mathbf{H}$. In the latter case, the estimation consists of prediction
and update steps, similarly to the classical Kalman filter. However,
we no longer update the prediction with an additive term but we have
a nonlinear update function.

\subsection{Relation to Other Works \label{sub:Relation-to-Other}}

The proposed state-space representation in (\ref{eq:state-space-rep}) provides a framework to exploit the
correlation between the consecutive columns of the nonnegative data matrix
in NMF. Several approaches have been proposed in the literature
to exploit the temporal dynamics in NMF, such as frame stacking \cite{Gemmeke2011a},
convolutive NMF \cite{Smaragdis2007,Wang2009}, smooth NMF \cite{Fevotte2009,Mohammadiha2011a,Mohammadiha2012a,Mohammadiha2013d},
and state-space representations \cite{Mysore2010,Nakano2010,Mohammadiha2013c,Mohammadiha2013,Mohammadiha2013b,Fevotte2013}. The state-space representations (including our proposed approach) model the interdependencies between different rows of the NMF coefficient matrix $\mathbf{H}$, unlike the smooth NMF approaches that assume these rows to be independent. Most of these approaches can be explained in a unified framework \cite{Smaragdis2014}. Our proposed approach is most related to the N-HMM approach in \cite{Mysore2010} and the nonnegative dynamical system (NDS) in \cite{Fevotte2013}.

Both our proposed D-NMF approach and the N-HMM approach in \cite{Mysore2010} use
the PLCA framework and provide a state-space representation to benefit
from the temporal dynamics. However, unlike the N-HMM approach that uses a
discrete state-space representation, our approach is based on a continuous state-space representation. The principal difference between both approaches is hence
the same as the difference between HMM and Kalman filter. A continuous
dynamical system is superior if the underlying source signal smoothly
transits between many (possibly infinite) states, whereas a discrete
dynamical system can be more suitable if the source signal switches between a limited number
of states. Hence, N-HMM can for example be a good model for speech if we
assume that a speech signal exhibits a switching behavior between limited number
of phonemes. On the other hand, a continuous state-space representation is more appropriate
for multitalker babble noise, since it is generated as the sum of a number of speech signals, and there are in principle many states obtained by
the combination of the states of the underlying speech signals. A thorough discussion on this example can be found in \cite{Mohammadiha2013c}.
Another important difference between our proposed D-NMF method and the N-HMM methods in \cite{Mysore2010,Mohammadiha2013c} is computational complexity. To analyze
a mixture of two (or more) sources where each source is individually
modeled using an N-HMM, a factorial N-HMM has to be used. This leads
to exponential complexity in the number of sources for N-HMM based
systems, and approximate inference approaches, e.g., \cite{Mysore2012},
have to be used to keep the complexity tractable. In contrast, the
complexity of the D-NMF approach is linear in the number of sources and no approximation is needed.

Similar to our D-NMF approach, the NDS approach in \cite{Fevotte2013} uses
a continuous state-space representation that is written as:
\begin{multline}
\left\{ \begin{array}{ll}
f\left(\mathbf{h}_{t}\mid\mathbf{A}_{1},\mathbf{h}_{t-1},\boldsymbol{\alpha}\right)=\text{v-gamma}\left(\mathbf{h}_{t};\boldsymbol{\alpha},\mathbf{A}_{1}\mathbf{h}_{t-1}/\boldsymbol{\alpha}\right),\\
f\left(\mathbf{x}_{t}\mid\mathbf{W},\mathbf{h}_{t},\delta\right)=\text{v-gamma}\left(\mathbf{x}_{t};\delta\boldsymbol{1},\mathbf{W}\mathbf{h}_{t}/\delta\right),
\end{array}\right.\label{eq:state-space-rep-nds}
\end{multline}
where $I$-dimensional vector $\boldsymbol{\alpha}$ and scalar
$\delta$ are model parameters, $\boldsymbol{1}$ is a $K$-dimensional
all-ones-vector, division of vectors is performed element by element,
and $\text{v-gamma}\left(\mathbf{h}_{t}\mid\boldsymbol{\alpha},\boldsymbol{\beta}\right)$
corresponds to a gamma distribution over the vector $\mathbf{h}_{t}$
with independent elements $h_{it}$ as
\begin{equation}
\text{v-gamma}\left(\mathbf{h}_{t}\mid\boldsymbol{\alpha},\boldsymbol{\beta}\right) = \prod_{i=1}^{I}\frac{1}{\beta_{i}^{\alpha_{i}}\Gamma\left(\alpha_{i}\right)}h_{it}^{\alpha_{i}-1}e^{-h_{it}/\beta_{i}},\label{eq:gamma_dist}
\end{equation}
where $\Gamma\left(\cdot\right)$ is the gamma function. Using (\ref{eq:gamma_dist}),
the conditional expected values of the state variables and data are given as $E(\mathbf{h}_{t}\mid\mathbf{A}_{1},\mathbf{h}_{t-1},\boldsymbol{\alpha})=\mathbf{A}_{1}\mathbf{h}_{t-1}$, and
$E(\mathbf{x}_{t}\mid\mathbf{W},\mathbf{h}_{t},\delta)=\mathbf{W}\mathbf{h}_{t}$, respectively.

There are three main differences
between our D-NMF and the NDS approaches. Firstly, the NDS approach assumes that each element of the
observation vector ($x_{kt}$) is a gamma-distributed random variable,
while in our approach the observation vectors ($\mathbf{x}_{t}$) are
multinomial-distributed. These assumptions lead to two different
NMF cost functions, where one of them may be preferred for a specific application \cite{Smaragdis2014}. The NDS method minimizes the IS divergence
and hence is a dynamical IS-NMF, while our method minimizes a weighted
KL divergence and hence is the dynamical counterpart for PLCA.
Additionally, the assumed distribution for the NMF coefficients corresponds to an exponential and a gamma distribution for the NDS approach and our D-NMF approach, respectively. Secondly, our proposed D-NMF approach provides a more general multi-lag
predictor for the state variables, while the NDS approach (as well as N-HMM approaches)
use a one-lag predictor, i.e., $J=1$. Thirdly, our proposed estimation
approach (Section \ref{sub:Estimation-Algorithm}) has appealing
properties regarding the estimation of the state variables and the
N-VAR model parameters. Our estimation of the state variables consists
of two steps corresponding to a prediction and an update step,
similar to a Kalman filter, which leads to an easy and intuitive explanation
of the update rules. Moreover, we show that the N-VAR model parameters
can be estimated by applying a separate NMF, which is more suitable
in the nonnegative framework. Neither of these properties are provided
in the NDS approach.
\subsection{Estimation Algorithm\label{sub:Estimation-Algorithm}}

In this section, we derive an EM algorithm to estimate the nonnegative parameters
in (\ref{eq:state-space-rep}), which are denoted by $\boldsymbol{\lambda}=\left\{ \mathbf{A},\mathbf{H},\mathbf{W}\right\} $,
given a nonnegative data matrix $\mathbf{X}$. We aim to maximize the
MAP objective function for the model given in (\ref{eq:state-space-rep}),
(\ref{eq:exp_prior}), and (\ref{eq:mult_dist}), i.e.,
as\footnote{Note that $f\left(\mathbf{h}_t\right)$ (as part of $f\left(\mathbf{H}\right)$)
  in \eqref{eq:MAP_objective} is not only conditioned on $\mathbf{A}$
  but also on ${\mathbf{h}}_{t-1},\ldots{\mathbf{h}}_{t-J}$. The latter conditioning is omitted in this equation to keep the notations uncluttered.}:
\begin{alignat}{1}
Q^{\text{MAP}} & =\log f\left(\mathbf{X},\mathbf{H}\mid\mathbf{W},\mathbf{A}\right)\nonumber \\
 & =\log f\left(\mathbf{X}\mid\mathbf{W},\mathbf{H}\right)+\log f\left(\mathbf{H}\mid\mathbf{A}\right).\label{eq:MAP_objective}
\end{alignat}

Maximizing $Q^{\text{MAP}}$ w.r.t. $\mathbf{W},\mathbf{A}$ and $\mathbf{H}$ results in a MAP estimate of $\mathbf{H}$
and ML estimates of $\mathbf{W}$ and $\mathbf{A}$. For this optimization,
we derive an EM algorithm \cite{bishop2006}, which is a commonly used
approach to estimate the unknown parameters in the presence of latent
variables. The EM algorithm maximizes a lower bound on $Q^{\text{MAP}}$ and iterates
between an expectation (E) step and a maximization (M) step until convergence.
We denote the EM latent variables by $z_t$, an indicator variable to
index the basis vectors. In the E step, the posterior probabilities
of these variables are obtained as:
\begin{eqnarray}
f\left(z_{t}=i\mid k,\boldsymbol{\lambda}\right) & = & \frac{f\left(k\mid z_t=i\right)f\left(z_{t}=i\right)}{\sum_{i=1}^{I}f\left(k\mid z_t=i\right)f\left(z_{t}=i\right)}\nonumber \\
 & = & \frac{w_{ki}h_{it}}{\sum_{i}w_{ki}h_{it}},\label{eq:E_step}
\end{eqnarray}
where $\boldsymbol{\lambda}$ denotes the estimated parameters from
the previous iteration of the EM algorithm. In the M step,
the expected log-likelihood of the complete data \cite[Chapter 9]{bishop2006}%
\footnote{For $t\leq J$, $\hat{\mathbf{h}}_{t-j}$
is set to a vector consisting of ones to prevent accessing undefined variables.%
}:
\begin{eqnarray}
Q\left(\hat{\boldsymbol{\lambda}},\boldsymbol{\lambda}\right) & = & \sum_{t,i}f\left(z_{t}=i\mid k,\boldsymbol{\lambda}\right)\log f\left(\mathbf{x}_{t},z_{t}\mid\hat{\boldsymbol{\lambda}}\right)\notag\\
 & + & \sum_{t}\log f\left(\hat{\mathbf{h}}_{t}\mid\hat{\mathbf{A}},\hat{\mathbf{h}}_{t-1},\ldots\hat{\mathbf{h}}_{t-J}\right),\label{eq:EM_lowerbound_prior}
\end{eqnarray}
is maximized w.r.t. $\hat{\boldsymbol{\lambda}}$ to obtain
a new set of estimates. Note that using Jensen's inequality, it can be easily proved that $Q\left(\hat{\boldsymbol{\lambda}},\boldsymbol{\lambda}\right)$
is a lower bound for $Q^{\text{MAP}}$. Using (\ref{eq:exp_prior}) and (\ref{eq:mult_dist}), $Q\left(\hat{\boldsymbol{\lambda}},\boldsymbol{\lambda}\right)$
can be equivalently (up to a constant) written as (also see \cite{Shashanka2007a}):
\begin{eqnarray}
Q\left(\hat{\boldsymbol{\lambda}},\boldsymbol{\lambda}\right) & = & \sum_{k,t,i}x_{kt}f\left(z_{t}=i\mid k,\boldsymbol{\lambda}\right)\left(\log\hat{w}_{ki}+\log\hat{h}_{it}\right)\notag\\
 & - & \sum_{i,t}\left(\log\hat{\eta}_{it}+\frac{\hat{h}_{it}}{\hat{\eta}_{it}}\right),\label{eq:EM_lowerbound2_prior}
\end{eqnarray}
where $\hat{\boldsymbol{\eta}}_{t}=\sum_{j=1}^{J}\hat{\mathbf{A}}_{j}\hat{\mathbf{h}}_{t-j}$.
As mentioned in Section \ref{sub:Statistical-Model-Description}, $\mathbf{w}_{i}$ and $\mathbf{h}_{t}$
are probability vectors, and hence, to make sure that they sum to one, we need to impose two constraints $\sum_{i}\hat{h}_{it}=1$ and $\sum_{k}\hat{w}_{ki}=1$. To solve the constrained optimization problem, we form the Lagrangian function $\mathcal{L}$ and maximize it:
\begin{equation}
\mathcal{L}\hspace{-.8mm}=\hspace{-.8mm}Q\left(\hat{\boldsymbol{\lambda}},\boldsymbol{\lambda}\right)+
\sum_{i}\alpha_{i}\left(1\hspace{-.8mm}-\hspace{-.8mm}\sum_{k}\hat{w}_{ki}\right)+\sum_{t}\beta_{t}\left(1\hspace{-.8mm}-\hspace{-.8mm}\sum_{i}\hat{h}_{it}\right),
\label{eq:complete_objective_function}
\end{equation}
where $\alpha_i,i=1\ldots I$ and $\beta_t,t=1\ldots T$ are Lagrange multipliers. In the following,
we describe the maximization w.r.t. $\mathbf{W}$, $\mathbf{H}$,
and $\mathbf{A}$, respectively.

Eq. (\ref{eq:complete_objective_function}) can be easily maximized
w.r.t. $\hat{w}_{ki}$ to obtain:
\begin{equation}
\hat{w}_{ki}=\frac{\sum_{t}x_{kt}f\left(z_{t}=i\mid k,\boldsymbol{\lambda}\right)}{\alpha_{i}},\label{eq:basis_estimate}
\end{equation}
where the Lagrange multiplier $\alpha_{i}=\sum_{t,k}x_{kt}f\left(z_{t}=i\mid k,\boldsymbol{\lambda}\right)$
to ensure that $\hat{\mathbf{w}}_{i}$ sums to one. For the estimation
of $\mathbf{H}$, we propose a recursive algorithm, i.e., we estimate
$\hat{\mathbf{h}}_{1},\hat{\mathbf{h}}_{2},\ldots$ sequentially.
Therefore, we first predict the state variables as
\begin{equation}
\hat{\mathbf{h}}_{t\mid t-1}=\hat{\boldsymbol{\eta}}_{t}=\sum_{j=1}^{J}\hat{\mathbf{A}}_{j}\hat{\mathbf{h}}_{t-j},\label{eq:predict_h}
\end{equation}
where $\hat{\mathbf{h}}_{t\mid t-1}$ is the  prediction result given all the past observations $\mathbf{x}_{1},\ldots\mathbf{x}_{t-1}$.
In the update step, the current observation $\mathbf{x}_{t}$ is used
to update the state estimate. This is done by maximizing (\ref{eq:complete_objective_function})
w.r.t. $\hat{\mathbf{h}}_{t}$. Setting the derivative of $\mathcal{L}$
w.r.t. $\hat{h}_{it}$ to zero, we obtain:
\begin{equation}
\hat{h}_{it\mid t}=\hat{h}_{it}=\frac{\sum_{k}x_{kt}f\left(z_{t}=i\mid k,\boldsymbol{\lambda}\right)}{\beta_{t}+1/\hat{\eta}_{it}}.\label{eq:update_h}
\end{equation}

The Lagrange multiplier $\beta_{t}$ has
to be computed such that $\hat{\mathbf{h}}_{t}$ sums to one, for which we have used an
iterative Newton\textquoteright{}s method.

Finally, the estimation of the N-VAR parameters $\hat{\mathbf{A}}$ is presented in the following.
Note that there are many approaches to estimate the VAR model parameters
in the literature \cite{Hamilton1994,Luetkepohl2005}. However, since
most of these approaches are based on least-squares estimation, they are
not suitable for our nonnegative framework. Moreover, they tend to be
very time-consuming for high-dimensional data. First, let us define
the $I\times IJ$-dimensional matrix $\hat{\mathbf{A}}$ as: $\hat{\mathbf{A}}=\left[\hat{\mathbf{A}}_{1}\;\hat{\mathbf{A}}_{2}\ldots\hat{\mathbf{A}}_{J}\right]$.
Accordingly, let $IJ$-dimensional vector $\hat{\mathbf{v}}_{t}$ represent the stacked state
variables as: $\hat{\mathbf{v}}_{t}^{\top}=\left[\hat{\mathbf{h}}_{t-1}^{\top}\;\hat{\mathbf{h}}_{t-2}^{\top}\ldots\hat{\mathbf{h}}_{t-J}^{\top}\right]$.
The parts of (\ref{eq:complete_objective_function}) that depend on $\hat{\mathbf{A}}$ are equivalently written as:
\begin{eqnarray}
\mathcal{L}^{\left(A\right)} & = & -\sum_{i,t}\left(\log\left[\hat{\mathbf{A}}\hat{\mathbf{v}}_{t}\right]_{i}+\frac{\hat{h}_{it}}{\left[\hat{\mathbf{A}}\hat{\mathbf{v}}_{t}\right]_{i}}\right)\nonumber \\
 & = & -d_{\text{IS}}\left(\mathbf{\hat{H}\|}\hat{\mathbf{A}}\mathbf{\hat{V}}\right)-\sum_{i,t}\left(\log\hat{h}_{it}+1\right),\label{eq:object_A}
\end{eqnarray}
where $\mathbf{\hat{V}}=[\hat{\mathbf{v}}_{1}\ldots \hat{\mathbf{v}}_{T}]$,  $\left[\cdot\right]_{i}$ denotes the $i$-th entry of its argument, and $d_{\text{IS}}\left(\cdot\|\cdot\right)$ is the IS divergence
as defined in (\ref{eq:IS_cost}). The second term in (\ref{eq:object_A})
is constant and can be ignored for the purpose of optimization w.r.t.
$\hat{\mathbf{A}}$. Hence, the ML estimate of $\hat{\mathbf{A}}$ can be obtained by performing IS-NMF in which the NMF coefficient
matrix $\mathbf{\hat{V}}$ is held fixed and only the basis matrix
$\hat{\mathbf{A}}$ is optimized. This is done by executing (\ref{eq:updateW_IS})
iteratively until convergence. Alternatively, we can repeat (\ref{eq:updateW_IS})
only once resulting in a generalized EM algorithm. We used
the latter alternative in our simulations.

The proposed estimation approach for the N-VAR parameters is able to automatically capture the importance weight for each lag, i.e., $\mathbf{A}_j,j=1\ldots J$  are not required to, e.g., be normalized to have the same $l_1$ norm. Hence, different lags may contribute differently, proportional to their norm, in computing $\sum\mathbf{A}_{j}\mathbf{h}_{t-j}$. This is achieved because the NMF coefficients $\mathbf{\hat{V}}$ are held fixed
in the IS-NMF, and we no longer have a scale ambiguity in the NMF representation.
%



\begin{algorithm}[t]
\caption{\label{alg:Algorithm1} Proposed dynamic NMF: algorithm to learn the model parameters.}
{\small
\begin{enumerate}
  \item Set the predefined variables $I$ (number of NMF basis vectors), $J$ (N-VAR model order), and $M,q$ (see the text).
  \item Initialize $\hat{\mathbf{W}},\hat{\mathbf{H}}$ and $\hat{\mathbf{A}}_j$ for $j=1 \ldots J$ with positive random numbers. Set $r=1$.
  \item Repeat until convergence:
  \begin{enumerate}
    \item Compute $\hat{\mathbf{W}}$  using \eqref{eq:basis_estimate}
    \item Compute the state variables $\hat{\mathbf{H}}$
\begin{algorithmic}[0]
\FOR{$t=1:T$}

    \STATE {\color{Gray} \% Predict}
    \IF{$r>M$}
    \STATE Compute $\hat{\boldsymbol{\eta}}_{t}$ using \eqref{eq:predict_h}.
    \STATE Anneal the prediction as $\hat{\eta}_{it}=\hat{\eta}_{it}^q$.
    \ELSE
    \STATE Set $\hat{\boldsymbol{\eta}}_{t}$ to all-ones-vector.
    \ENDIF
    \STATE {\color{Gray} \% Update}
    \STATE Update the state estimate $\hat{\mathbf{h}}_{t}$ using \eqref{eq:update_h}.
\ENDFOR
\end{algorithmic}
\item Compute the N-VAR parameters
\begin{algorithmic}[0]
\IF{$r\geq M$}
\STATE Compute $\hat{\mathbf{A}}_j$ for $j=1 \ldots J$ using \eqref{eq:object_A} and \eqref{eq:updateW_IS}.
\ENDIF
\end{algorithmic}
\item $r=r+1.$
  \end{enumerate}
  \end{enumerate}
  }
\end{algorithm}
Algorithm \ref{alg:Algorithm1} summarizes our proposed D-NMF approach to estimate all the model parameters
simultaneously, which is usually applied on the training data (c.f. Section \ref{sec:Numerical-Simulations}) to learn the model parameters  $\mathbf{W}$ and $\mathbf{A}$. As convergence criterion, the stationarity of $Q^{\text{MAP}}$ or EM lower bound can be checked, or a fixed (sufficient) number of iterations can be simply used. In our simulations, we
have used 100 iterations. This algorithm includes two practical
additions. First, since the EM algorithm converges to a local optimum of the
objective function, a good initialization can improve the performance.
Therefore, we have introduced a parameter $M$ that is used to postpone the
estimation of $\hat{\mathbf{A}}$ until a relatively good ML estimate
of the state variables $\hat{\mathbf{H}}$  has been found. We intuitively set $M$ to half of the maximum number of iterations ($M=50$).
Additionally, we have defined a parameter $q$ that is used to anneal (or weight) the predictions, and it was experimentally set to 0.15 in our experiments. Intuitively, this heuristic trick takes into account the uncertainties
(as the covariances in the Kalman filtering), and it was found to be beneficial in our simulations.

Algorithm \ref{alg:Algorithm2} summarizes our filtering algorithm where the model parameters
(including $\mathbf{W}$ and $\mathbf{A}$) are learned a priori and
held fixed during the process, as it is done in classical Kalman
filtering. Here, motivated by the simulated annealing, we use an adaptive annealing of the predictions. Intuitively, the predictions are effectively used in the first iterations to prevent the EM algorithm to get stuck in a local maximum. Then, the predictions are smoothed over the iterations causing the NMF approximation to be a better fit
to the current observation.
Moreover, for practical problems where the dynamics of
unseen data can never be learned accurately, this adaptive annealing
makes the algorithm more robust.

\begin{algorithm}[t]
\caption{\label{alg:Algorithm2} Proposed dynamic NMF: filtering algorithm applied at time $t$.}
{\small
\begin{enumerate}
  \item Set the predefined variable $q<1$
  \item Initialize $\hat{\mathbf{h}}_t$ with positive random numbers. Load the model parameters $\mathbf{W}$ and $\mathbf{A}$ learned using Algorithm \ref{alg:Algorithm1}. Set $r=1$.
  \newline
  {\color{Gray} \% Predict}
  \item Compute predictions:
\begin{enumerate}
    \item Compute $\hat{\boldsymbol{\eta}}_{t}$ using \eqref{eq:predict_h}.
    \item Backup the prediction $\mathbf{b}_t=\hat{\boldsymbol{\eta}}_{t}$.
  \end{enumerate}
  {\color{Gray} \% Update}
  \item Repeat until convergence:
  \begin{enumerate}
      \item Anneal the prediction as $\hat{\eta}_{it}=b_{it}^{q/r}$.
      \item Update the state estimate $\hat{\mathbf{h}}_{t}$ according to \eqref{eq:update_h}.
      \item $r=r+1.$
  \end{enumerate}
\end{enumerate}
}
\end{algorithm}

\section{Numerical Simulations\label{sec:Numerical-Simulations}}

In this section, we present our experimental results using the proposed D-NMF algorithm. We have performed simulations for three examples, namely tracking the frequency of a single sinusoid in noise (Section \ref{sub:Tracking-the-Frequency}), separation of two signals with similar basis (Section \ref{sub:Separation}), and speech denoising (Section \ref{sub:Denoising}). Since the original time-domain signals in the described examples can take negative values, we need to transform them to a nonnegative domain. For this purpose, we apply a discrete Fourier transformation (DFT) to Hann-windowed (overlapping) short-time frames to obtain a complex-valued time-frequency representation of the input signals. We then use the magnitudes of the DFT coefficients to construct the nonnegative observation matrix
to be used with NMF. We compare the performance
of the proposed D-NMF approach using
objective measures with the performance of the static NMF approach \cite{Smaragdis2006} and two other NMF approaches that exploit the temporal dynamics, namely the N-HMM approach in \cite{Mysore2010} and a frame stacking approach \cite{Gemmeke2011a}. The signal-to-noise ratio (SNR) is used to quantify the noise level in the observations.
Denoting the clean (not known to the algorithms) and noisy time-domain
signals as $\mathsf{x}$ and $\mathsf{y}$, the input SNR is defined as:
\begin{equation}
\text{Input SNR}=10\log_{10}\frac{\sum_{n}\mathsf{x}_{n}^{2}}{\sum_{n}\left(\mathsf{y}_{n}-\mathsf{x}_{n}\right)^{2}},\label{eq:input_SNR}
\end{equation}
where $n$ is the sample index.

\subsection{Tracking the Frequency of a Single Sinusoid in Noise\label{sub:Tracking-the-Frequency}}

\begin{figure}
\includegraphics[width=.95\columnwidth]{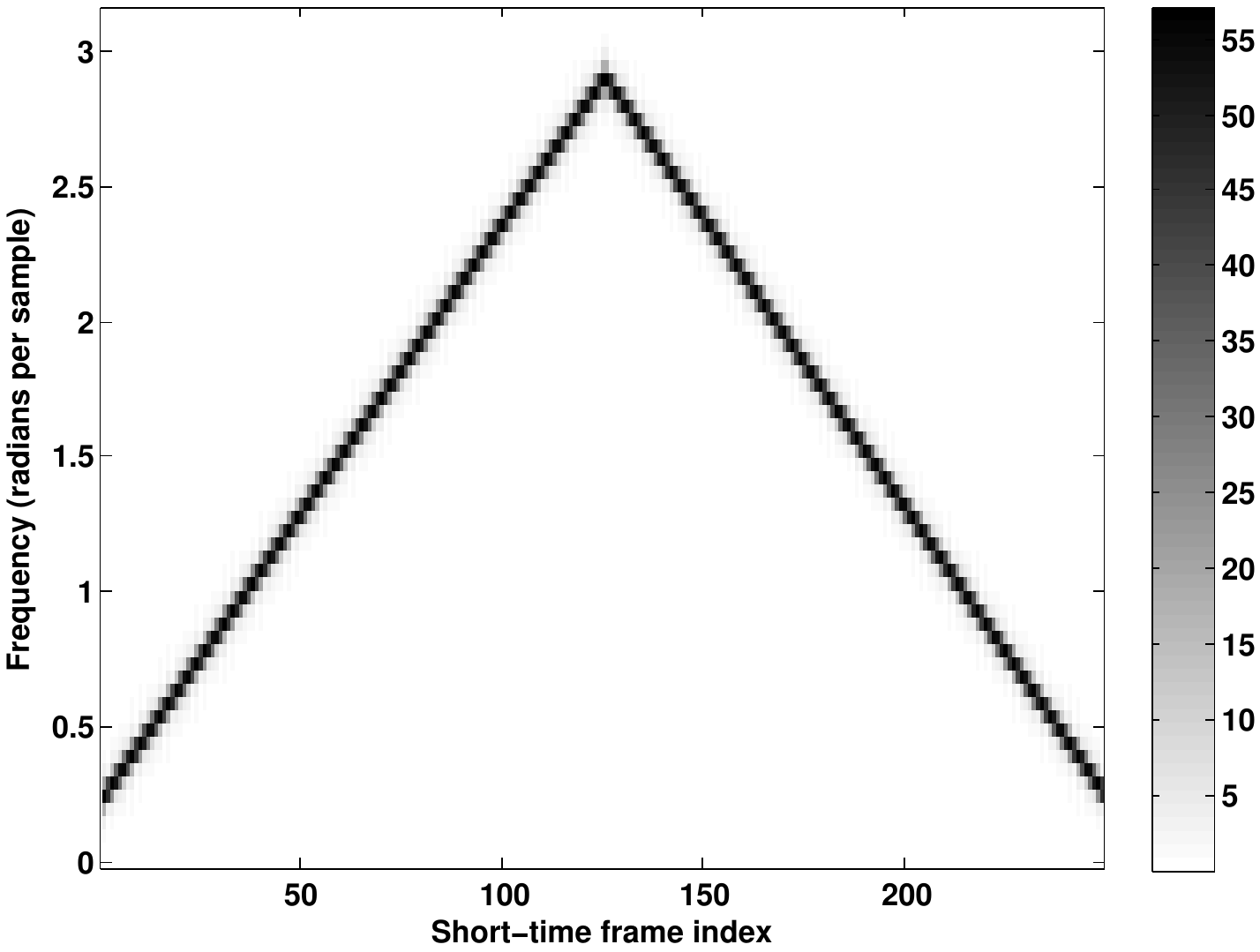}\caption{\label{fig:sinusoid} Time-frequency representation of the DFT magnitudes of a single
sinusoid with time-varying frequency. First the frequency increases, and
before reaching the Nyquist frequency (at the 127-th frame) it gradually reduces.}
\end{figure}

In this section, the performance of the proposed D-NMF approach is demonstrated
using a tracking example. Estimation of the frequency and phase
of sinusoids in noise is still an active area of research \cite{Fu2007}.
In this experiment, we aim to estimate the frequency of a single
sinusoid in the presence of noise with high levels. The target signal is sampled at a sampling frequency of 8 kHz. The frequency of the sinusoid is time-varying, and increases from 0.24 radians/sample (300 Hz) to 2.9 radians/sample (3700 Hz) and then reduces to 0.24 radians/sample again. The DFT with a frame length of 128 samples and a (non-overlapping) Hann window was applied and the obtained magnitude spectrogram was used as the nonnegative observation matrix%
\footnote{Although the DFT results in a $128\times T$-dimensional matrix,
because of the symmetry property of the DFT for real-valued signals,
we only use the first $K=65$ rows as the observation matrix.%
}. Fig.~\ref{fig:sinusoid} depicts the noise-free observation matrix.
Here, the $k$-th element of $\mathbf{x}_{t}$ is proportional to the signal\textquoteright{}s
energy at a specific frequency given by $2\pi\left(k-1\right)/128$
radians/sample.

For the simulations, white Gaussian noise was added to the target signal at various input SNRs. In the NMF approaches, the basis matrix $\mathbf{W}$ was predefined (and was held fixed) as the
identity matrix of size $65\times65$. We set $J=1$, and since we
do not expect any large jump of the frequency, we predefined
$\mathbf{A}_{1}$ such that the diagonal elements and their adjacent
neighbors have a value of $1/3$ while the rest of the elements are
set to zero. This assumption means that the frequency will either stay constant or will smoothly increase or decrease to a higher or a lower value, respectively.

To estimate the frequency in each short-time frame, NMF
or D-NMF ($J=1,q=0.25$ in Algorithm \ref{alg:Algorithm2}) was first applied and then the frequency was computed as
$\hat{\Omega}_{t}=2\pi\left(k_{\text{max}}-1\right)/128$ radians/sample
in which $k_{\text{max}}$ is the index of the maximum entry of $\mathbf{h}_{t}.$ For comparison purposes, the frequency was also estimated using an N-HMM approach. The N-HMM consisted of 65 states with one spectral vector per state. The same basis matrix $\mathbf{W}$ and $\mathbf{A}_{1}$ were used to predefine the N-HMM state spectral vectors and transition matrix. This N-HMM is effectively an HMM where the state-conditional likelihoods are computed using a multinomial distribution. For the N-HMM approach, $k_{\text{max}}$ is the index of the state with the highest posterior probability, which is determined by applying the forward algorithm \cite{Rabiner1989}.

\begin{figure}
\includegraphics[width=.95\columnwidth]{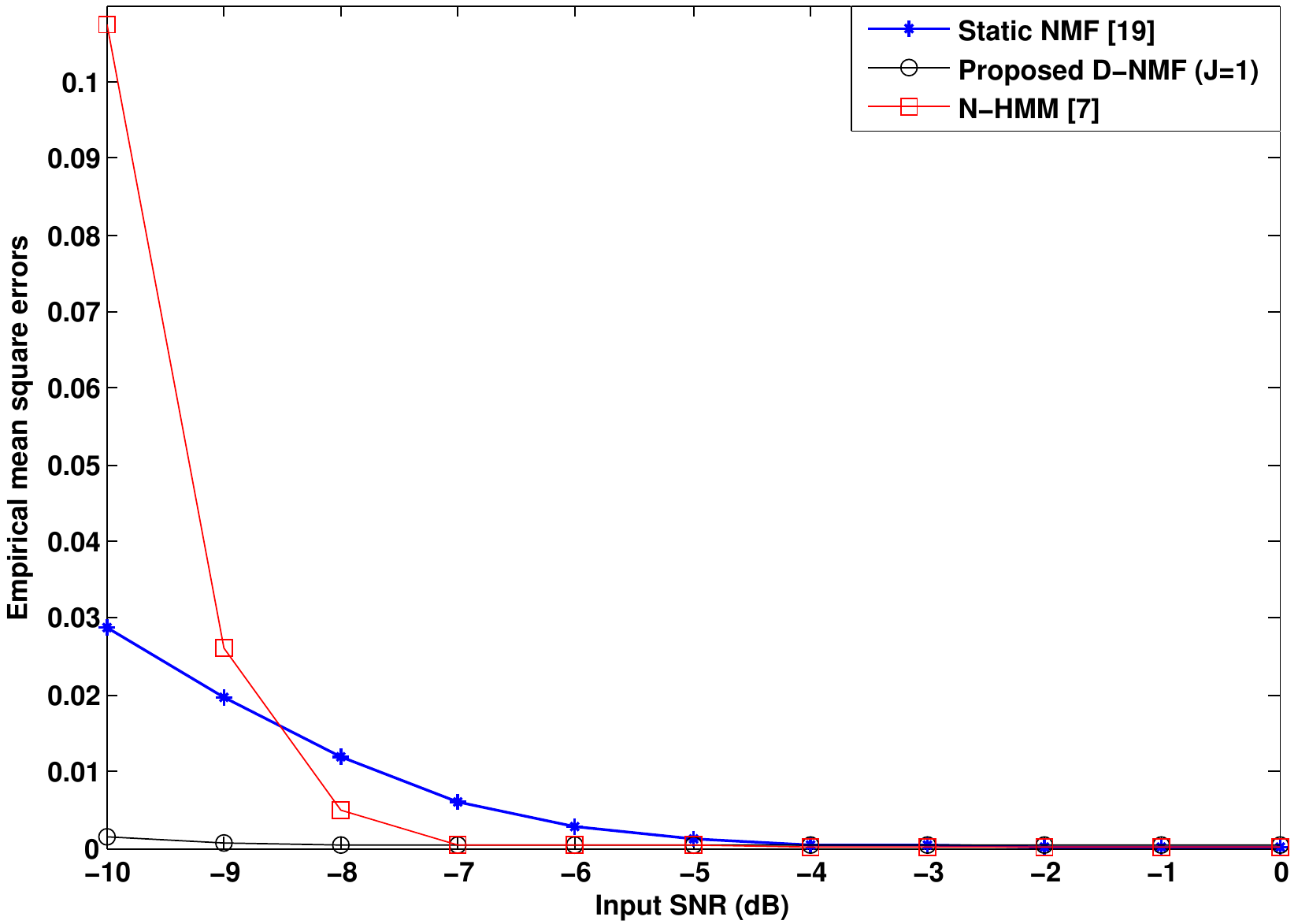}\caption{\label{fig:Empirical-MSE} Empirical mean square errors in tracking
the frequency of a single sinusoid as a function of the input SNR.}
\end{figure}

The tracking performance is evaluated using the empirical mean square
error:
\begin{equation}
\text{MSE}=\frac{1}{LT}\sum_{l=1}^{L}\sum_{t=1}^{T}\left(\hat{\Omega}_{t}-\Omega_{t}\right)^{2},
\end{equation}
where $\Omega_{t}$ is the ground-truth frequency, and $L$ is the number
of Monte Carlo runs that is set to $50$ in our simulations. Fig.~\ref{fig:Empirical-MSE}
shows the MSE as a function of the input SNR. As can be seen, the D-NMF
approach provides a significantly smaller error compared to the static NMF and N-HMM approaches, especially at low input SNRs. The performance of the N-HMM approach degrades quickly at low input SNRs, which indicates that the approach is not as robust as D-NMF to high noise levels. The difference arises from the fact that in the N-HMM approach, the state-conditional likelihoods are used during the forward algorithm, which are sensitive to high noise levels and exhibit a large dynamic range. For the D-NMF approach, however, the posterior probabilities $f\left(z_{t}=i\mid k,\boldsymbol{\lambda}\right)$ that are used to compute $\hat{\mathbf{h}}_{t}$ (in \eqref{eq:update_h}) have a smaller dynamic range and the noise effect can be more effectively compensated by using the temporal continuity.
For higher input SNRs, the input data matrix exhibits a
clearer energy distribution (closer to the noise-free case) such
that applying the NMF and the N-HMM approaches will also lead to good results. The simulation results shows
that at an input SNR of about -3 dB, applying D-NMF leads to slightly
larger error than static NMF. This error is due to the additional latency that is
imposed by using the previous observations to predict the current
state variables.

\subsection{Separation of Two Signals with Similar Basis\label{sub:Separation}}

\begin{figure}
\includegraphics[width=.95\columnwidth]{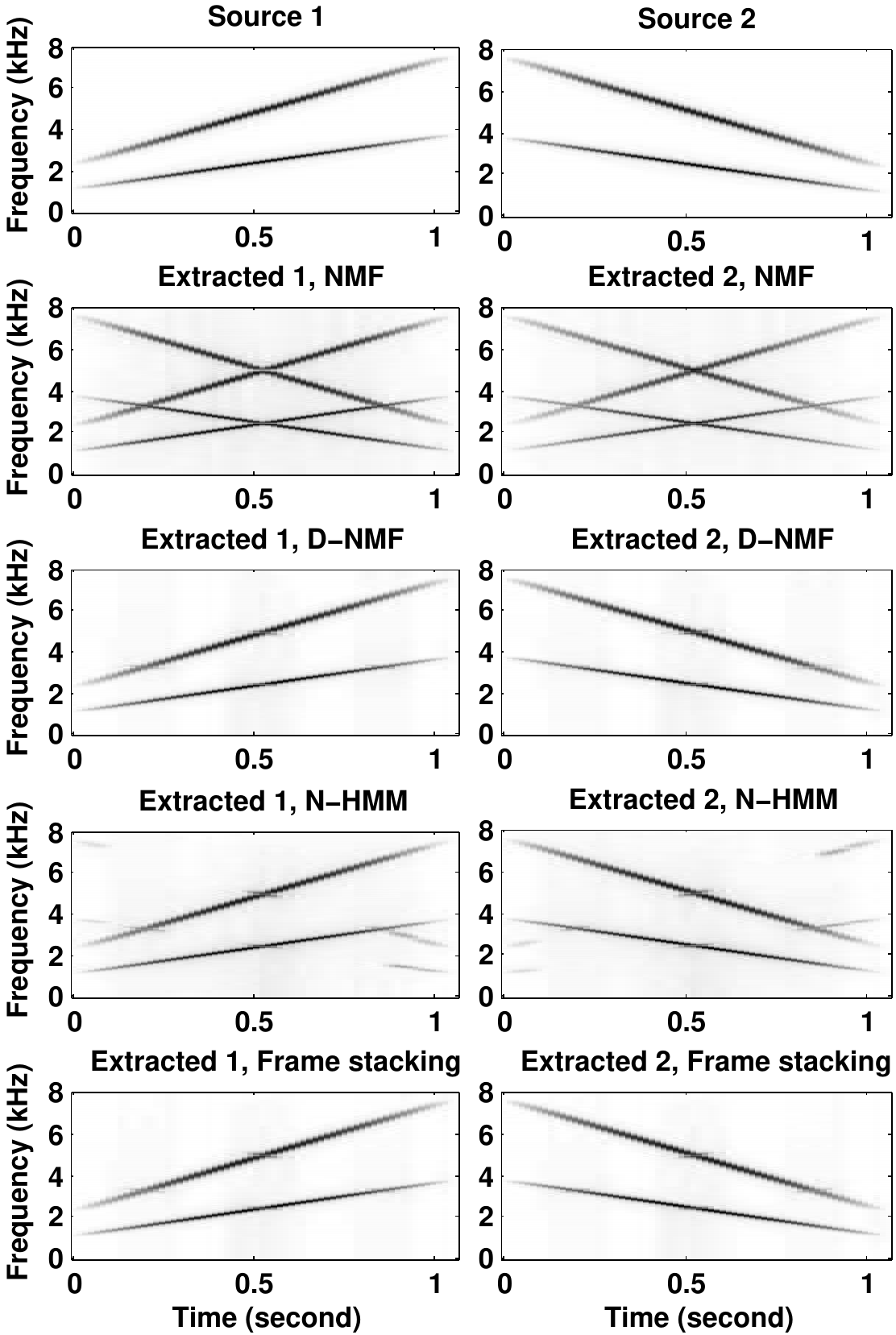}\caption{\label{fig:bss} Original and separated sources using NMF, D-NMF ($J=2$), N-HMM, and frame stacking approaches.}
\end{figure}

In the second experiment, we applied our proposed D-NMF approach as a supervised separation approach for separating two
sources that share a similar basis matrix $\mathbf{W}$. In this experiment, two sources
(each consisting of two sinusoids with time-varying frequencies) were added at an input SNR of 0 dB to obtain the time-domain mixture. The DFT was applied using overlapping Hann windows with a frame length of 1024 samples and an update length
of 256 samples. The sampling frequency was 16 kHz. The magnitude spectrogram of the two sources are separately shown
in the top panel of Fig.~\ref{fig:bss}. Although these sources share
a similar basis matrix (because they are just time-reversed
versions of each other) they have a very different dynamic behavior.
The frequencies of source 1 are increasing, while the frequencies of source 2 are decreasing.

To learn the model parameters, the static NMF and D-NMF approaches were applied on the observations of each source separately. The number of basis vectors was set to $I=50$ for each source both for NMF and for D-NMF. For the D-NMF approach, in addition to the basis matrix $\mathbf{W}$, the N-VAR model parameters $\mathbf{A}_j,j=1\ldots J$ were learned for $J\in\{1,2,3,4,5\}$. The annealing parameter $q$ was set to 0.1 in Algorithm \ref{alg:Algorithm2}. In addition to the static NMF and the proposed D-NMF approaches, an N-HMM approach \cite{Mysore2010} and a frame stacking approach \cite{Gemmeke2011a} were implemented as alternative methods that exploit the temporal dynamics in NMF. For the N-HMM approach, 50 states with one spectral vector per state were learned for each source. For the frame stacking approach, 8 consecutive frames were stacked to obtain 4096-dimensional vectors and a tall basis matrix with $I=50$ columns were learned to represent each source. In this experiment and also in Section \ref{sub:Denoising}, the high-resolution DFT-domain magnitude spectral vectors are stacked rather than the low-resolution mel-domain spectral vectors which is proposed in \cite{Gemmeke2011a} because the DFT version outperformed the mel counterpart. The number of N-HMM states, the number of basis vectors, and the number of consecutive frames to be stacked were experimentally set to get the best performance.

To model the mixture, we
assume that the DFT magnitude of the mixture is (approximately) equal to the sum of the magnitudes of the DFT coefficients of the two sources \cite{Smaragdis2007,Mysore2010,Mohammadiha2013d},
i.e., $\mathbf{x}_{t}\approx\mathbf{x}_{t}^{\left(1\right)}+\mathbf{x}_{t}^{\left(2\right)}$,
where the superscripts represent the source numbers. For the NMF, D-NMF and frame stacking approaches, the basis matrix of the mixture is constructed by concatenating
the (learned) individual basis matrices, i.e, $\mathbf{W}=\left[\begin{array}{cc}
\mathbf{W}^{\left(1\right)} & \mathbf{W}^{\left(2\right)}\end{array}\right]$. Similarly, for the D-NMF approach, the N-VAR parameters of the mixture are constructed by concatenating the (learned) individual N-VAR parameters, i.e., $\mathbf{A}_{1}=\left[\begin{array}{ccccc}
\mathbf{A}_{1}^{\left(1\right)} & \mathbf{0} & ; & \mathbf{0} & \mathbf{A}_{1}^{\left(2\right)}\end{array}\right]$ where $\mathbf{0}$ is a $50\times50$ zero matrix.  For the N-HMM approach, a factorial N-HMM \cite{Mysore2010} is constructed to model the mixture.

For the separation, the basis matrix $\mathbf{W}$ and N-VAR parameters
$\mathbf{A}_{j}$ are held fixed and only the NMF coefficients
$\mathbf{H}={[\begin{array}{cc}
\mathbf{H}^{(1),\top} & \mathbf{H}^{(2),\top}\end{array}]}^\top$ are estimated. For all the approaches, after convergence of the estimation algorithm, the magnitude of the DFT coefficients of the individual
sources are estimated using a Wiener reconstruction \cite{Mohammadiha2013b}:
\begin{equation}
\mathbf{\hat{x}}_{t}^{\left(1\right)}=\frac{\mathbf{W}^{\left(1\right)}\mathbf{h}_{t}^{\left(1\right)}}{\mathbf{W}^{\left(1\right)}\mathbf{h}_{t}^{\left(1\right)}+\mathbf{W}^{\left(2\right)}\mathbf{h}_{t}^{\left(2\right)}}\odot\mathbf{x}_{t},
\label{eq:wiener_reconstruction}
\end{equation}
where $\odot$ represents the element-wise multiplication, and division is
performed element by element. The separated signals using the NMF,
D-NMF ($J=2$), N-HMM, and frame stacking approaches are shown in Fig.~\ref{fig:bss}. As can be seen, the static NMF approach is not able to separate the sources because of the ambiguity that is caused by the similarity of the individual
basis matrices. On the other hand, the three other approaches, lead to a satisfactory
separation of the sources by benefiting from the temporal dependencies, where the D-NMF and frame stacking approaches have clearly led to a better separation compared to the N-HMM approach.

To quantify the separation performance, the output SNR was computed as:
\begin{equation}
\text{Output SNR}=10\log_{10}\frac{\sum_{n}\mathsf{x}_{n}^{2}}{\sum_{n}\left(\hat{\mathsf{x}}_{n}-\mathsf{x}_{n}\right)^{2}},\label{eq:output_SNR}
\end{equation}
where $\mathsf{x}$ is the time-domain signal corresponding to one
of the sources, and $\hat{\mathsf{x}}$ is the separated time-domain signal, obtained by applying the overlap-add procedure
to the separated magnitude spectrogram, where the phase of the mixture signal was used to compute the inverse DFT.

\begin{figure}
\includegraphics[width=.95\columnwidth]{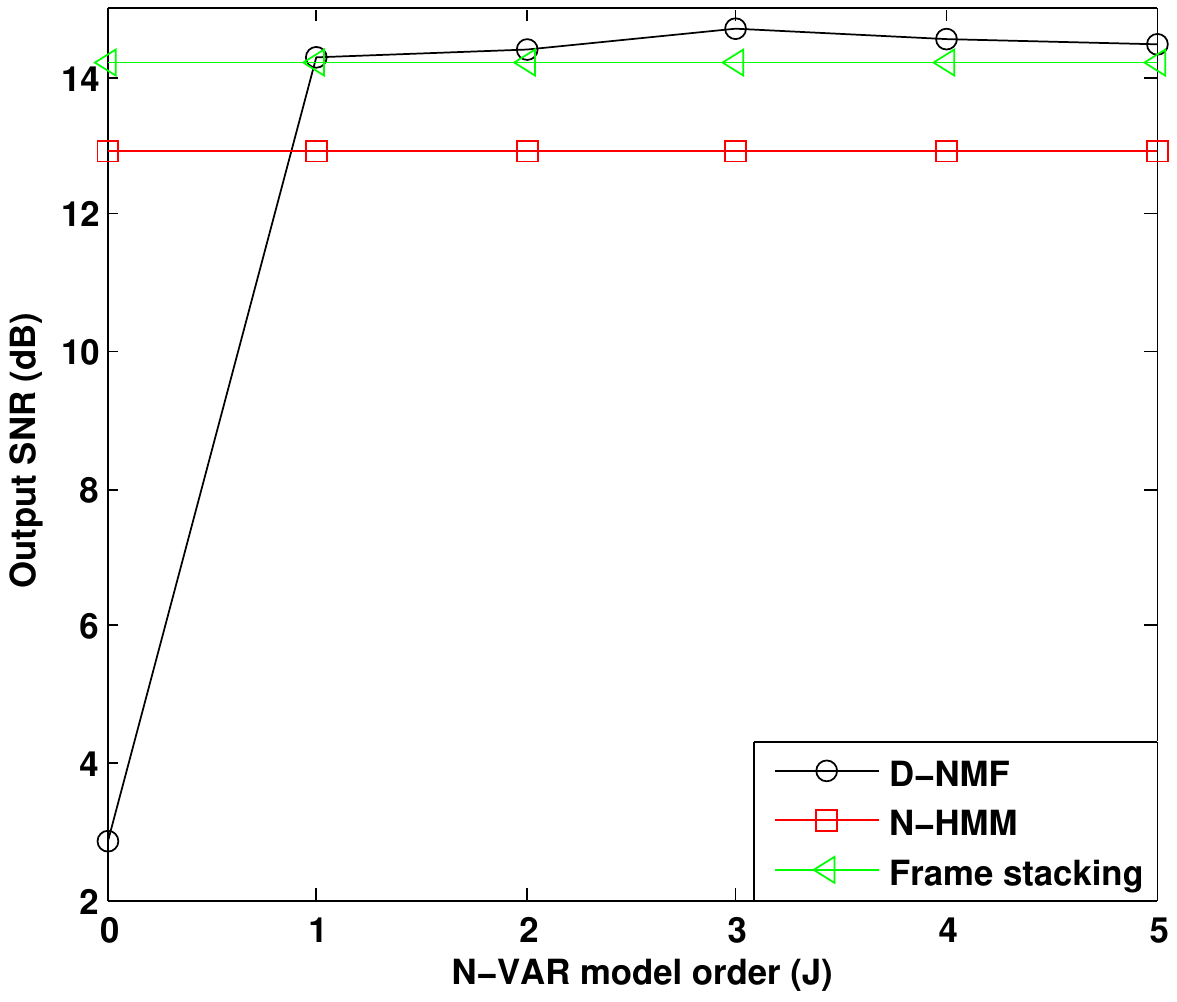}\caption{\label{fig:bss_order} Output SNR as a function of the N-VAR model order $J$. $J=0$ corresponds to the static NMF approach with no temporal modeling.}
\end{figure}

Fig.~\ref{fig:bss_order} shows the output SNR as a function of the N-VAR model order ($J$). Here, $J=0$ corresponds to the static NMF approach with no temporal modeling. As can be seen, including temporal dynamics in NMF has improved the output SNR by more than 11 dB. By increasing $J$, the performance slightly improves, reaching its maximum at $J=3$ for this experiment. Moreover, as also shown in Fig.~\ref{fig:bss}, the D-NMF and frame stacking approaches have produced higher output SNRs compared to the N-HMM approach, where D-NMF has led to the best separation performance.

\subsection{Denoising\label{sub:Denoising}}
As the last experiment, we applied our proposed D-NMF approach to a speech
denoising problem. In this experiment, speech signals are degraded
by additive noise and the goal is to suppress the noise and estimate
the speech component given the noisy observations. The speech signals were degraded with multitalker babble noise
or factory noise at input SNRs in the range -5 dB to 5 dB. The speech and noise signals were taken from the TIMIT \cite{Garofolo1993} and NOISEX-92 \cite{noisex92} databases, respectively. The signals were
sampled at a sampling frequency of 16 kHz. The DFT analysis was performed with the same parameters as in Section \ref{sub:Separation}.

For each noise type, an NMF model was learned using the first 75\% of the noise signals and the last 25\% was used to test the algorithms. The noise type is assumed to be known to choose a suitable noise-dependent NMF model for denoising. This assumption is practical for some applications and the required information can be provided by state-of-the-art environment classification techniques (see \cite{Mohammadiha2013d} for a discussion on this topic). The denoising was performed under two conditions, depending on the available information about the speaker identity. In the matched condition, the speaker identity is assumed to be known and speaker-dependent (SD) speech models were used in all the approaches. These models were learned using 9 speech sentences from each speaker, and another sentence from the same speaker was used to test the algorithms. Alternatively, in the mismatched case, a universal speaker-independent (SI) speech model was learned using 200 speech sentences from different speakers. The denoising experiments were repeated for 20 different speakers, where the training and test data were disjoint in all the simulations. For all the methods, the speech DFT magnitudes were estimated using the Wiener reconstruction \eqref{eq:wiener_reconstruction}.

The number of basis vectors for speech and noise were experimentally found for each approach to obtain the best results. For the NMF, D-NMF, and frame stacking approaches, $I=60$ speech basis vectors were learned for both SD and SI models, where for the N-HMM approach, 40 and 60 states each consisting of 10 spectral vectors were respectively learned for the SD and SI models. For the NMF and D-NMF approaches, 20 basis vectors were learned for each noise type, where for the frame stacking approach, 100 and 150 basis vectors were learned for babble and factory noise, respectively. For the N-HMM approach a single-state model was learned for each noise type, where the number of spectral vectors was set to 20 (for both noise types in the SD condition) and to 20 and 100 (in SI condition) for babble and factory noise types, respectively. For the D-NMF approach, the N-VAR model parameters were learned for $J\in\{1,2,3,4,5\}$, where the annealing parameter $q$ was experimentally set to 0.3 (for speech) and 0.1 (for both noise types) in Algorithm \ref{alg:Algorithm2}.

\begin{figure}
\includegraphics[width=.95\columnwidth]{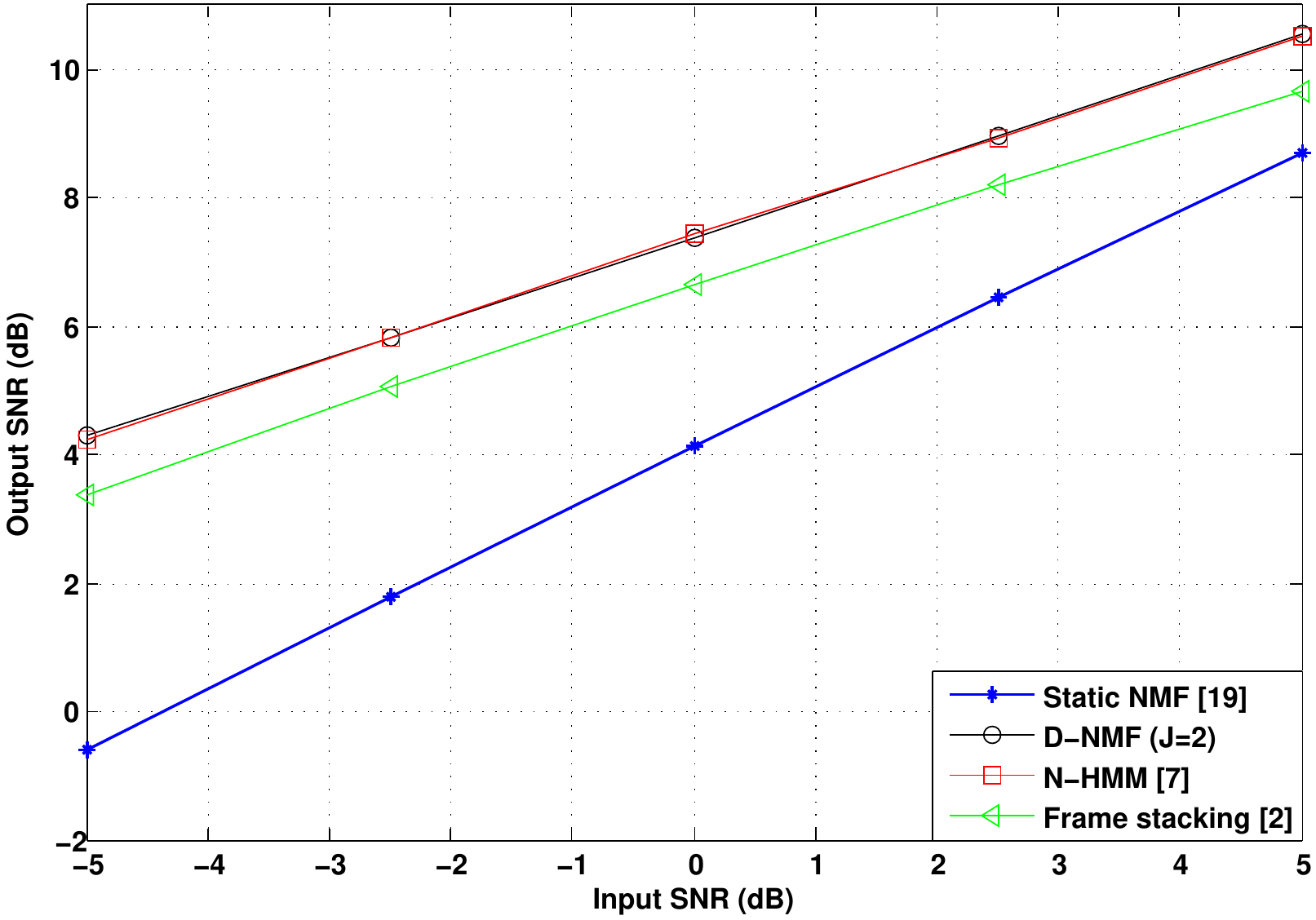}
\caption{\label{fig:Results-of-denoisingSD} Averaged output SNR over the factory and babble noise types under the matched speaker-dependent condition. Noise type and speaker identity are assumed to be known a priori and they are used to select noise- and speaker-dependent models for denoising.}
\end{figure}

Fig.~\ref{fig:Results-of-denoisingSD} shows the results (averaged over both noise types) of our denoising
experiment for the matched SD condition (with $J=2$ for the D-NMF approach). The figure shows the output SNR, defined in (\ref{eq:output_SNR}),
as a function of the input SNR in the range of -5 dB to 5 dB. The simulation
results show that the D-NMF and N-HMM approaches have a similar denoising performance, while they significantly outperform the static NMF approach for
all considered input SNRs. The difference is maximum at the lowest input SNR (-5 dB), where the D-NMF approach results in around 4.5 dB higher output SNR. Moreover, the frame stacking approach has a considerably improved performance compared to the NMF approach, but is worse than the D-NMF and N-HMM approaches.

\begin{figure}
\includegraphics[width=.95\columnwidth]{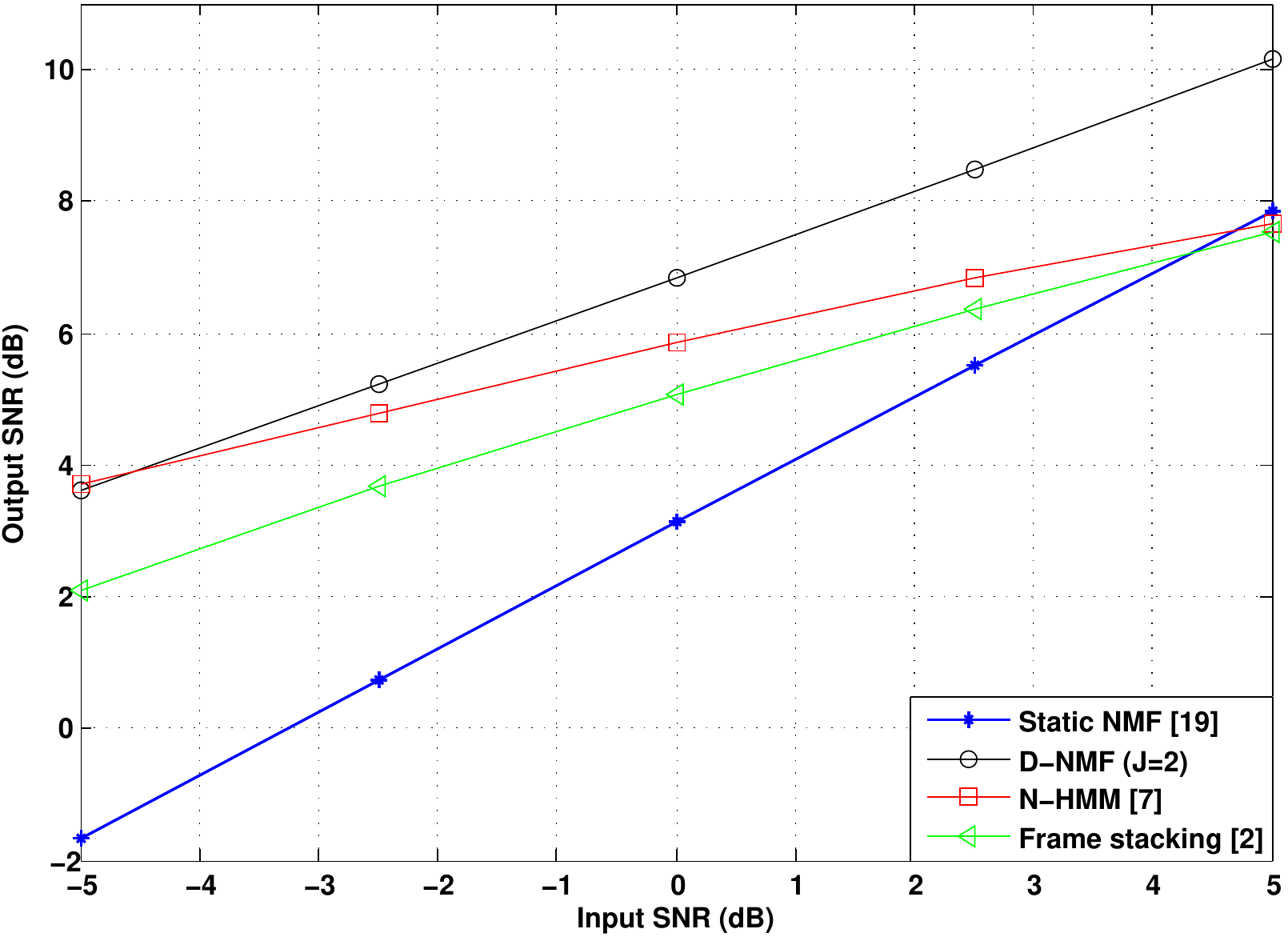}
\caption{\label{fig:Results-of-denoisingSI} Averaged output SNR over the factory and babble noise types, where a universal speaker-independent speech model is used for denoising. Noise type is assumed to be known a priori and is used to select a noise-specific NMF (or N-HMM) model for denoising.}
\end{figure}

The results of the denoising approaches (averaged over both noise types) for the mismatched SI condition are shown in Fig.~\ref{fig:Results-of-denoisingSI}. The results show that the D-NMF ($J=2$) approach outperforms both the N-HMM and frame stacking approaches, where the difference is more than 2 dB at an input SNR equal to 5 dB. Comparing Fig.~\ref{fig:Results-of-denoisingSD} and Fig.~\ref{fig:Results-of-denoisingSI}, we see that a higher input SNR is obtained under the SD condition.

\begin{figure}
\includegraphics[width=.95\columnwidth]{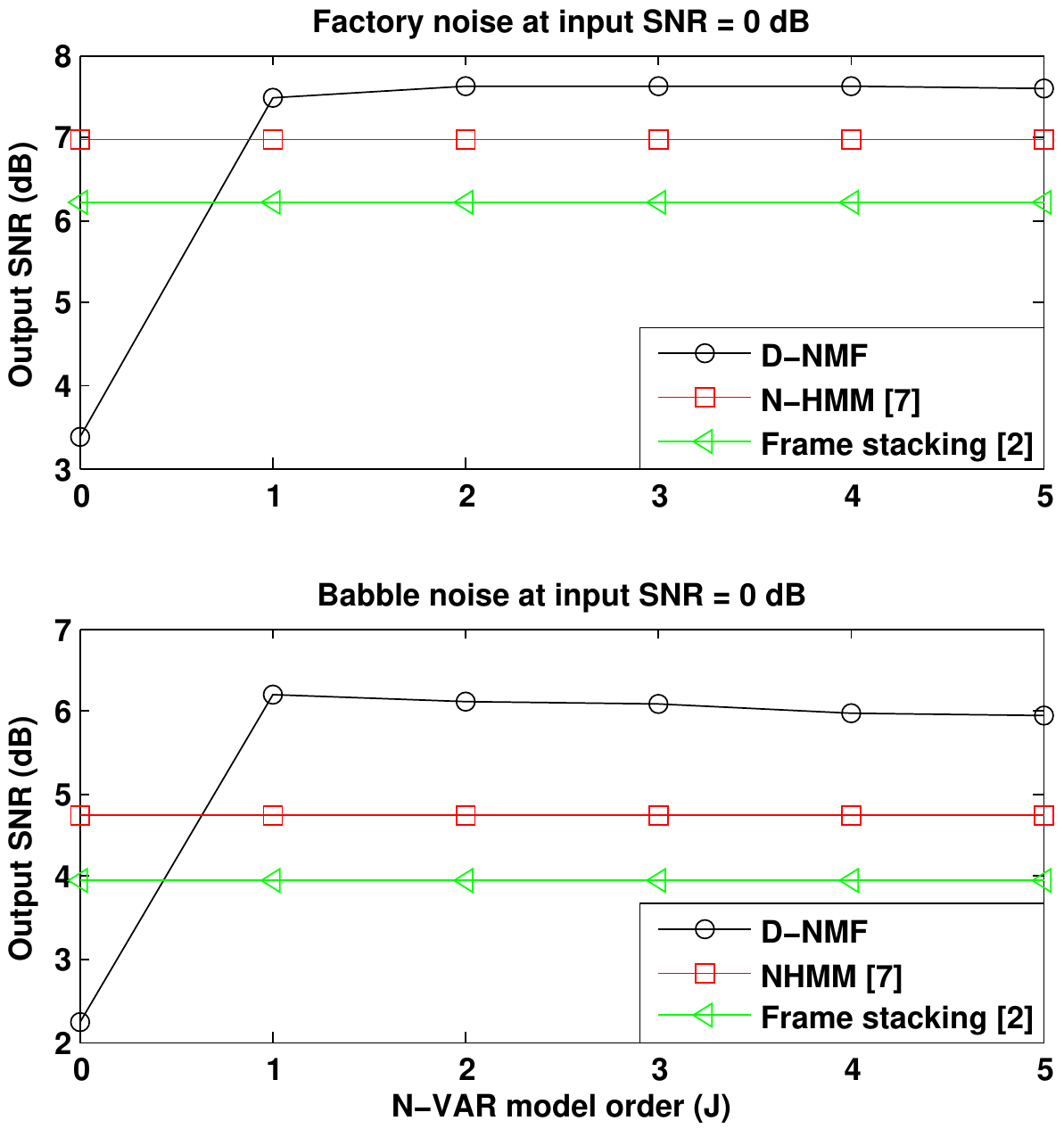}
\caption{\label{fig:scan_order_denoising} Output SNR corresponding to the factory noise (top panel) and babble noise (bottom panel) as a function of
the N-VAR model order $J$ at input SNR = 0 dB. Universal speaker-independent speech model is used for denoising.}
\end{figure}

Fig.~\ref{fig:scan_order_denoising} shows the output SNR as a function of the N-VAR model order $J$, for the SI condition and at an input SNR equal to 0 dB. The results for factory noise and babble noise types are plotted in the top and bottom panels, respectively. The static NMF is shown as a special case of D-NMF with $J=0$. The results show that a significant improvement is obtained by incorporating the temporal dynamics into the denoising process. For the factory noise, a small improvement is obtained by increasing $J$ to 3, while for the babble noise the best performance is obtained at $J=1$. In both cases, it can be seen that a single-lag predictor with $J=1$ can be used to achieve a good denoising performance.

Finally, it is interesting to compare the computational complexity and the memory requirement of the proposed D-NMF approach to the N-HMM and frame stacking approaches. To have a better understanding, we simply provide an estimate of the required time to process one second of speech in the SD denoising example in our implementation in a PC
with 3.4 GHz Intel CPU and 8 GB RAM. It should be mentioned that this time can be significantly reduced for all the approaches by using an optimized implementation. Our D-NMF approach requires around 1.5 seconds to process 1 second of input signal, while the N-HMM and frame stacking approaches require 40 and 0.75 seconds, respectively. Considering the memory requirements (to store the learned model parameters), D-NMF requires less than 25\% and 10\% of the memory required by the N-HMM and frame stacking approaches, respectively. As a result, the proposed approach is more suitable for real-time applications with power or memory restrictions.

\section{Conclusions\label{sec:Conclusions}}
In this paper, we presented a state-space representation for nonnegative
observations and nonnegative state variables, which is able to efficiently model the temporal dependencies. Since the classical
Kalman filtering is not appropriate for this setup, we derived a novel
algorithm, referred to as D-NMF, to learn the model parameters, and we developed a novel filtering approach for the state variables. Using an iterative EM-based estimation algorithm, an ML estimate of the basis matrix and the N-VAR model
parameters is computed. We showed that computing the ML estimate of the N-VAR parameters
is equivalent to applying IS-NMF in which the observations and the NMF
coefficients are the estimates of the state variables and their shifted versions, respectively.
As for the state variables, the algorithm provides a MAP solution
that, similar to the Kalman filtering, consists of a prediction step and
an update step. We demonstrated the algorithm using three examples
targeting tracking, separation, and denoising applications. The results
show that exploiting the temporal dynamics in NMF can improve the performance significantly,
especially at low input SNRs. Moreover, our experimental results show that the proposed approach outperforms an N-HMM and a frame stacking approach where it also requires substantially less computational power and memory, and hence, it is a better alternative for real-time applications. Finally, our approach to model the temporal dependencies is causal, i.e., it only uses the past observations to process the current observation. Therefore, unlike the frame stacking approach that has an inherent delay of several time steps, our approach does not impose any delay on the processed signals.
\ifCLASSOPTIONcaptionsoff
  \newpage
\fi
\bibliographystyle{plain}
{
\bibliography{NasserRefs}
}


\end{document}